\documentclass{article}

\usepackage{microtype}
\usepackage{graphicx}
\usepackage{subfigure}
\usepackage{booktabs} 
\usepackage{url}
\usepackage{footmisc}
\usepackage{amsmath}
\usepackage{amssymb}
\usepackage{amsfonts}
\usepackage{multirow}

\usepackage{hyperref}

\usepackage[accepted]{icml2021}

\icmltitlerunning{DRTCI}

\begin{document}

\twocolumn[
\icmltitle{DRTCI: Learning Disentangled Representations for Temporal Causal Inference}



\icmlsetsymbol{equal}{*}
\begin{icmlauthorlist}
\icmlauthor{Garima Gupta}{tcs}
\icmlauthor{Lovekesh Vig}{tcs}
\icmlauthor{Gautam Shroff}{tcs}
\end{icmlauthorlist}

\icmlaffiliation{tcs}{TCS Research, Delhi, India}

\icmlcorrespondingauthor{Garima Gupta}{gupta.garima1@tcs.com}

\icmlkeywords{Machine Learning, ICML}

\vskip 0.3in
]




\printAffiliationsAndNotice{}

\begin{abstract}
Medical professionals evaluating alternative treatment plans for a patient often encounter time varying confounders, or covariates that affect both the future treatment assignment and the patient outcome. The recently proposed Counterfactual Recurrent Network (CRN) accounts for time varying confounders by using adversarial training to balance recurrent historical representations of patient data. However, this work assumes that all time varying covariates are confounding and thus attempts to balance the full state representation. 
Given that the actual subset of covariates that may in fact be confounding is in general unknown, recent work on counterfactual evaluation in the static, non-temporal setting has suggested that disentangling the covariate representation into separate factors, where each either influence treatment selection, patient outcome or both can help isolate selection bias and restrict balancing efforts to factors that influence outcome, allowing the remaining factors which predict treatment without needlessly being balanced.
We hypothesize that such disentanglement should be possible in the temporal setting as well, and would be beneficial when dealing with time varying confounders. We propose  DRTCI, a model for temporal causal inference 
which uses a recurrent neural network to learn hidden representation of the patient's evolving covariates that disentangles into three factors that each causally determine either treatment, outcome or both treatment and outcome.
The model is evaluated on the same simulated model of tumour growth used to evaluate the CRN, with varying degrees of time-dependent confounding. %
The resulting outcome predictions from DRTCI significantly outperform the predictions from existing baselines 
especially for cases with high confounding and minimal historical data (early prediction). Ablation experiments are additionally performed to identify the key contributing factors to the performance of DRTCI. 
\end{abstract}

\section{Introduction and Motivation}
\label{sec:intro}
Critical medical decisions pertaining to treatment plan selection require reasoning about potential future outcomes given a patient's current state and observed longitudinal clinical data. Standard randomized control trials, the most reliable technique for evaluating different treatment options, are often impractical in the temporal setting, as it may not be feasible to conduct a trial for every temporal variation of a particular treatment plan. 

Thus there is an acute need to estimate the effects of different future treatment sequences over time from observational patient data. However, a major hurdle in this endeavour is the presence of time-varying confounders, or patient covariates that are causally influenced by past treatments and also influence future treatments and outcomes \cite{cole2010illustrating}. Consider a Covid-19 patient having low fever (time varying confounder) in the early stages of the infection. The patient is then  administered steroids which spikes the fever and prompts administration of anti-virals,  but the patient does not survive. Without accounting for the time varying confounder (fever), one may reach the incorrect conclusion that anti-virals are harmful to covid patients.

\begin{figure}[!b]
  \centering
  \includegraphics[scale=0.3]{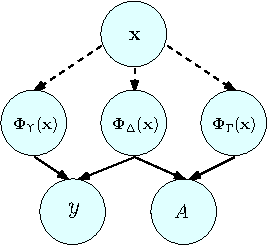}
  \caption{Disentanglement of static covariates ($\mathbf{x}$) into factors $\mathbf{\Phi}_{\Gamma}(\mathbf{x})$, $\mathbf{\Phi}_{\Upsilon}(\mathbf{x})$ and $\mathbf{\Phi}_{\Delta}(\mathbf{x})$  determining treatment $A$, outcome ($Y$)  or both respectively \cite{hassanpour2019learning}}
\label{fig:disentangle}
\end{figure}

A notable and closely related attempt to address the problem of time-varying confounders for evaluation of treatment plans is the recently proposed Counterfactual Recurrent Network (CRN) \cite{bica2020} which uses a GRU based recurrent model to predict potential future treatment effects by forcing hidden representations $\mathbf{\Phi}(\cdot)$ that can balance out time-varying confounding. However, these balanced representations curtail the representation of covariates which influence both future treatment assignment and outcome prediction. To mitigate this, we take inspiration from the work in \cite{hassanpour2019learning} (Figure 1)  and disentangle the temporal representation into the factors that exclusively determine future treatment ($\mathbf{\Phi}_{\Gamma}(\cdot)$), response ($\mathbf{\Phi}_{\Upsilon}(\cdot)$) or both treatment and response ($\mathbf{\Phi}_{\Delta}(\cdot)$). Extending  the work by \cite{hassanpour2019counterfactual} to learn disentangled, temporal representations,  we propose Disentangled Representations for Temporal Causal Inference (DRTCI), a novel sequence-to-sequence architecture based technique to evaluate potential treatment plans. DRTCI's novelty lies in the disentanglement of the latent temporal representation into three factors which influence either treatment assignment, response or both (confounding). The resulting factorization prevents needless balancing of factors that solely influence treatment assignment, and results in improved response prediction over full state balancing \cite{bica2020}, especially when the degree of confounding is high. 




The disentanglement is more effective for predicting future treatment outcomes, as only the relevant factors $\mathbf{\Phi}_{\Delta}(\cdot)$ and $\mathbf{\Phi}_{\Upsilon}(\cdot)$  of the representation are being used to predict the outcome, and the selection bias is isolated to the factors $\mathbf{\Phi}_{\Gamma}(\cdot)$ and 
$\mathbf{\Phi}_{\Delta}(\cdot)$ while balancing via adversarial loss is restricted to $\mathbf{\Phi}_{\Upsilon}$. It can also be argued that balancing can be employed on all factors, however since balancing is time dependent, the assumption of temporal variation of treatment and that of non confounders being correlated might not hold. We apply balancing selectively in contrast to the CRN which balances the entire representation  including the $\mathbf{\Phi}_{\Gamma}(\cdot)$ factor which has no influence on the future outcome and  more crucially $\mathbf{\Phi}_{\Delta}(\cdot)$ which  represents the confounding factors between treatment and future outcome $y$. Instead, we demonstrate improved prediction by retaining the confounding factors $\mathbf{\Phi}_{\Delta}(\cdot)$ for outcome prediction and mitigating  the resulting selection  bias via importance weights that are learnt by conditioning treatment prediction solely on the $\mathbf{\Phi}_{\Delta}(\cdot)$ factor via an independent arm of the model.

Our contributions in this paper are as follows:
1)  We propose a novel recurrent model DRTCI to predict future treatment effects in a temporal setting with time-varying confounding
2)  We demonstrate that disentangling the recurrent hidden representations in DRTCI yields substantial improvements to the prediction of future treatment effects over the recently proposed CRN model, the current state of the art  
3) We further show that the disentanglement is especially effective in situations with  extreme confounding and is also beneficial for early prediction with less historical patient data 
4) We conduct ablation studies to determine which components of DRTCI are contributing most to the improved performance and provide an analysis of the same.


\section{Related Work}
\label{sec:literature}
 While the field of causal inference has made significant strides towards counterfactual estimation, solutions have largely been focused on settings  where the treatment and outcome depend only on a static snapshot of the covariates. These include statistical techniques that employ  re-weighting of the data instances to balance the source and target
distributions \cite{austin2011introduction}, \cite{bottou2013counterfactual}, \cite{swaminathan2015self} which while unbiased, are known to suffer from high variance.
The recent application  of representation learning techniques \cite{shalit2017estimating}, \cite{johansson2016learning}, have shown promise especially for settings with high cardinality of treatments \cite{sharma2020hi}, and continuous treatment dosages but these too are not applicable to temporal treatment settings with time-varying confounding and can not capture the effects of time-varying exposure to treatments. An additional regularization was proposed to augment the representation learning by using propensity scores to ensure that data points ($\mathbf{x}$) that were nearby in the data space are also nearby in the representation space $(\mathbf{\Phi}(\mathbf{x}))$ i.e. by matching  $P(A|\mathbf{x})$ and $P(A|\mathbf{\Phi}(\mathbf{x}))$, where  $A$ is the treatment \cite{sharma2020multimbnn}, \cite{yao2018representation}. However these methods suffer when there are factors in the data that only determine treatment and do not influence the outcome as these factors will affect the similarity matching when ideally they should be excluded prior to the matching  \cite{hassanpour2019counterfactual}.

A recent model that is closely related to this paper is the work by \cite{hassanpour2019learning} which demonstrated the benefits of learning disentangled representations for counterfactual regression (see Figure 1). The authors argue that learning a single representation that removes the selection bias completely is not ideal, as their may be confounding factors that influence not only treatment selection but also outcome. Therefore, many of the prior proposed loss functions for representation learning \cite{shalit2017estimating} had to play a delicate balancing act of removing selection bias, yet retaining sufficient predictive information in the representations to make accurate counterfactual predictions. 
Notably, while this approach led to superior counterfactual estimation, it was again restricted to non-temporal setting. 

 Recent attempts at addressing causal inference in the temporal setting advocate utilizing recurrent neural networks to model the effects of time-varying exposures. This includes the work by \cite{lim2018forecasting} who utilize a sequence-to-sequence model to estimate the inverse probability of treatment weights (IPTW) for improving Marginal Structural Models (MSMs) \cite{robins2000marginal}, \cite{mansournia2017handling} and the more recent CRN model described in the previous section.

\section{DRTCI: Preliminaries and Loss Functions}
\label{sec:dis}
We describe the problem of estimating the response for factual or counterfactual treatments in a longitudinal setting. Further, we highlight the major components and loss functions employed by our proposed approach  for temporal causal inference. We make the standard assumptions made by \cite{robins2000marginal} to estimate treatment effects i.e. consistency, positivity and no
hidden confounders (sequential strong ignorability) \cite{pearl2018book}.
\subsection{Problem Setting}
Consider an observational dataset $\mathcal{D}=\{\{\mathbf{x}^{i}_t, \mathbf{a}_t^i,y^{i}_{t+1} \}_{t=1}^{T^{i}} \cup \{\mathbf{v}^i\} \}_{i=1}^N$ for $N$ patients containing time dependent patient covariates $\mathbf{x}^{i}_t$ and static covariates $\mathbf{v}^i$ for each patient $i$. Each patient receives one of $K$ treatments from the available treatment options in the set $\{A_1,..,A_k,...A_K\}$ at each timestep. The treatment received at each timestep is denoted by a one hot vector $\mathbf{a}_t^i = [a_t^i(1),.., a_t^i(k),..,a_t^i(K)]$ where $a_t^i(k) \in \{0,1\}$. A response $y^{i}_{t+1}$ is observed at each timestep and is appended as part of the covariates $\mathbf{x}^i_{t+1}$. The patient superscript $i$ is omitted in the text later for simplicity.

Let a course of $\tau$ treatments be received from timestep $t$ to $t+\tau-1$ as $\mathbf{\bar{a}}_{t,\tau}=[\mathbf{a}_{t},\mathbf{a}_{t+1},..,\mathbf{a}_{t+\tau-1}]$ with history $\mathbf{h}_t$ and current covariates $\mathbf{x}_t$ available at timestep $t$. The patient history at timestep $t$ is represented by a sequence of the patient's time-varying covariates $\mathbf{\bar{x}}_{1,t-1}=[\mathbf{x}_1,...,\mathbf{x}_{t-1}]$, sequence of treatments $\mathbf{\bar{a}}_{1,t-1}=[\mathbf{a}_1,...,\mathbf{a}_{t-1}]$ and static covariates $\mathbf{v}$ as $\mathbf{h}_t = [\mathbf{\bar{x}}_{1,t-1}, \mathbf{\bar{a}}_{1,t-1},\mathbf{v}]$. We are interested in estimating the response $y_{t+\tau}$ for a course of treatments $\mathbf{\bar{a}}_{t,\tau}$ received at timestep $t$ with history $\mathbf{h}_t$ and covariates $\mathbf{x}_t$ as observants as in Eq. \ref{eq:estimand}:
\vspace{-0.9em}
\begin{equation}
    \mathbb{E}[{y}_{t+\tau}[\mathbf{\bar{a}}_{t,\tau}]|[\mathbf{h}_t,\mathbf{x}_t]]
    \label{eq:estimand}
\end{equation}
An impediment in the estimation of $y_{t+\tau}$ from observational data is the presence of bias introduced during treatment assignment. Covariates $\mathbf{x}_t$ and history $\mathbf{h}_t$ at timestep $t$ causally affect treatment assignment $\mathbf{\bar{a}}_{t,\tau}$, and are thus time-varying confounding in nature. 
To compensate for the effect of time-varying confounders for unbiased causal inference in the longitudinal scenario, we propose a (DRTCI) model for learning disentangled representation for temporal causal inference described below.

\subsection{Recurrent Network for State Representation}

In DRTCI, we employ a recurrent neural network to map the observant comprising of the (variable-length) patient history $\mathbf{h}_t$ and  covariates at timestep $t$ to a latent, fixed length state representation $\mathbf{s}_t$:

\begin{equation}
    \mathbf{s}_t = f_{t,s}([\mathbf{h}_t,\mathbf{x}_t])
    \label{eq:state}
\end{equation}
We propose to learn three disentangled representations of the latent state at each timestep. These representations are: 1) Outcome Representation ($\mathbf{\Phi}_{\Upsilon}(\mathbf{s}_t)$) which affects future  outcome $y_{t+1}$, 2) Confounding Representation ($\mathbf{\Phi}_{\Delta}(\mathbf{s}_t)$) which affects both future treatment assignment $\mathbf{a}_{t}$ and outcome $y_{t+1}$, and 3) Treatment Representation ($\mathbf{\Phi}_{\Gamma}(\mathbf{s}_t)$) which affects treatment assignment $\mathbf{a}_{t}$, are learned in a disengaged manner as discussed next.

\subsection{Disentangled Representation Learning}
These disentangled representations are learned by passing the state at each timestep through three disentangled, two-layered ELU Multi-Layer Perceptrons (MLPs): $\mathbf{\Phi}_{\Upsilon}(\cdot), \mathbf{\Phi}_{\Delta}(\cdot), \mathbf{\Phi}_{\Gamma}(\cdot)$, parameters of which are learned using a set of four loss functions namely Prediction Loss, Treatment Loss, Imbalance Loss, and Weighting Loss:

\textbf{Prediction Loss \& Weighting Loss}:
The concatentation of confounding, outcome representation and treatment at timestep $t$ is passed to a regression arm ($\mathbf{W}_{\Upsilon}$) comprising of MLP-ELU-MLP layers to predict outcome $\hat{y}_{t+1}$.
\begin{equation}
    \hat{y}_{t+1} = \mathbf{W}_{\Upsilon}([\mathbf{\Phi}_{\Upsilon}(\mathbf{s}_t),\mathbf{\Phi}_{\Delta}(\mathbf{s}_t),\mathbf{a}_t])
    \label{eq:nextStepOutcome}
\end{equation}
We use squared error loss for learning the parameters of confounding and outcome representations. However, since the objective is to predict factual as well as counterfactual outcome while the observational dataset contains only factual outcomes, we use an importance weighting of squared error to emphasize those instances that are more useful for the prediction of counterfactual outcomes. This importance weight is an extension of binary treatment importance weighting introduced in \cite{hassanpour2019counterfactual} to $K$ treatments:
\begin{equation}
    \small
    \omega_t = \sum_{k=1}^K \frac{p(\mathbf{\Phi}_{\Delta}(\mathbf{s}_t)|A_k)}{p(\mathbf{\Phi}_{\Delta}(\mathbf{s}_t)|\mathbf{a}_t)} = \sum_{k=1}^K \frac{p(\mathbf{a}_t)}{p(A_k)}\frac{p(A_k|\mathbf{\Phi}_{\Delta}(\mathbf{s}_t))}{p(\mathbf{a}_t|\mathbf{\Phi}_{\Delta}(\mathbf{s}_t))} 
    \label{eq:importanceweights}
\end{equation}
In Eq. \ref{eq:importanceweights}, $p(\mathbf{a}_t), p(A_k)$ are the marginal probabilities for factual treatment $\mathbf{a}_t$ and treatment option $A_k$ respectively and are computed from the observational dataset. Further, propensity $p(A_k|\mathbf{\Phi}_{\Delta}(\mathbf{s}_t))$ \footref{footnote:shorthand} is obtained by passing $\mathbf{\Phi}_{\Delta}(\mathbf{s}_t)$ via an $\mathbf{W}_{\Delta}(\cdot)$ (MLP and a softmax layer) with $\mathcal{L}_{t,W}$ weighting loss as: $ \mathcal{L}_{t,W}(\mathbf{W}_\Delta,\mathbf{\Phi}_{\Delta}) = - \sum_{k=1}^{K} a_t(k) log (p(A_k|\mathbf{\Phi}_{\Delta}(\mathbf{s}_t)))$. 
The weighted square error is then used as outcome regression loss ($\mathcal{L}_{t,P}$) as:  $\mathcal{L}_{t,P}(\mathbf{\Phi}_{\Delta},\mathbf{\Phi}_{\Upsilon},\mathbf{W}_\Upsilon) = \omega_t (y_{t+1} - \hat{y}_{t+1})^2$

\textbf{Treatment Loss}:
Treatment loss is employed for learning the logging policy that guides treatment assignment. For this, both the confounding ($\mathbf{\Phi}_{\Delta}(\mathbf{s}_t)$) and treatment ($\mathbf{\Phi}_{\Gamma}(\mathbf{s}_t)$) representations are passed through $\mathbf{W}_{\Gamma}(\cdot)$ (MLP and a softmax layer) to learn the propensity of treatment $A_k$ as $p(A_k|[\mathbf{\Phi}_{\Delta}(\mathbf{s}_t),\mathbf{\Phi}_{\Gamma}(\mathbf{s}_t)])$\footref{footnote:shorthand} $\forall k$. Treatment loss ($\mathcal{L}_{t,T}$) is then used for learning the weights of confounding and treatment representations as:
\vspace{-0.5em}
\begin{equation} 
    \resizebox{0.9\hsize}{!}{$\mathcal{L}_{t,T}(\mathbf{\Phi}_{\Delta},\mathbf{\Phi}_{\Gamma},\mathbf{W}_\Gamma) = - \sum_{k=1}^{K} a_t(k) log (p(A_k|[\mathbf{\Phi}_{\Delta}(\mathbf{s}_t),\mathbf{\Phi}_{\Gamma}(\mathbf{s}_t)]))$}
    \label{eq:treatmentLoss}
\end{equation}

\textbf{Imbalance Loss}:
To overcome the bias due to time-varying confounders, we use an adversarial imbalance loss ($\mathcal{L}_{t,I}$): $\mathcal{L}_{t,I}(\mathbf{\Phi}_{\Upsilon}) =  -\sum_{k=1}^{K} a_t(k) log (p(A_k|\mathbf{\Phi}_{\Upsilon}(\mathbf{s}_t)))$
which ensures that the outcome representation ($\mathbf{\Phi}_{\Upsilon}(\mathbf{s}_t)$) is not predictive of treatment assignment, i.e, $p(\mathbf{\Phi}_{\Upsilon}(\mathbf{s}_t)|A_1) = ... = p(\mathbf{\Phi}_{\Upsilon}(\mathbf{s}_t)|A_K)$. 
We compute propensity $p(A_k|\mathbf{\Phi}_{\Upsilon}(\mathbf{s}_t))$\footnote{We use shorthand $p_1(\mathbf{s}_t),p_2(\mathbf{s}_t),p_3(\mathbf{s}_t)$ for propensities $p(A_k|\mathbf{\Phi}_{\Delta}(\mathbf{s}_t)),p(A_k|[\mathbf{\Phi}_{\Delta}(\mathbf{s}_t),\mathbf{\Phi}_{\Gamma}(\mathbf{s}_t)]),p(A_k|\mathbf{\Phi}_{\Upsilon}(\mathbf{s}_t))$ in the text that follows. \label{footnote:shorthand}} by passing outcome representation via randomly initialized, fixed weight one-layer MLP followed by softmax. 

\begin{figure*}
  \centering
  \includegraphics[scale=0.5]{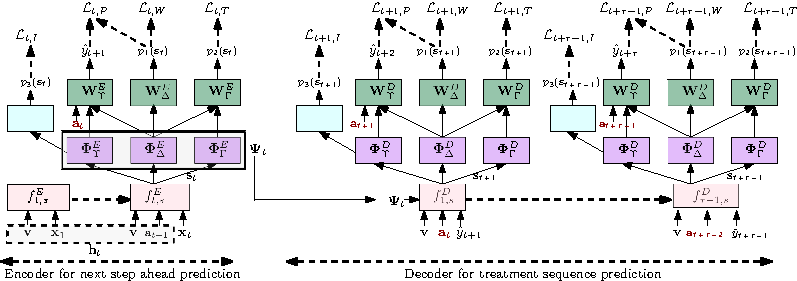}
  \caption{Figure illustrates architecture for DRTCI with encoder which builds disentangled representations that are unified as $\mathbf{\Psi}_t$ to initialize decoder which updates continuously to predict factual/counterfactual outcome to a sequence of treatments.}
\label{fig:architecture}
\end{figure*}

We shall describe the usage of loss functions for obtaining disentangled state representations in the proposed architecture of DRTCI in Section \ref{sec:DRTCI_arc} and Figure \ref{fig:architecture}.

\section{DRTCI: Proposed Architecture}
\label{sec:DRTCI_arc}
In this section, we describe the architecture and procedure for predicting outcome $y_{t+\tau}$ (Eq. \ref{eq:estimand}). We use a sequence-to-sequence architecture as described in \cite{bica2020,lim2018forecasting} with the added novelty of using disentangled state representations in the longitudinal setting. The encoder and decoder network of sequence-to-sequence architecture is discussed in the text that follows:
\subsection{Encoder}
The encoder network uses a recurrent network with an LSTM unit to process the observants comprising of history $\mathbf{h}_t$ and current covariates $\mathbf{x}_t$ to obtain state representation $\mathbf{s}_t$ (Eq. \ref{eq:state}). Further, the state representation is disentangled into an outcome representation ($\mathbf{\Phi}_{\Upsilon}^{E}(\mathbf{s}_t)$), a confounding representation ($\mathbf{\Phi}_{\Delta}^{E}(\mathbf{s}_t)$) and a treatment representation ($\mathbf{\Phi}_{\Gamma}^E(\mathbf{s}_t)$) using the loss at timestep $t$ as 
\begin{equation}
\small
\begin{split}
    \mathcal{L}_t(\mathbf{\Phi}_{\Delta}^E,\mathbf{\Phi}_{\Upsilon}^E,\mathbf{\Phi}_{\Gamma}^E,\mathbf{W}_{\Delta}^E,\mathbf{W}_{\Gamma}^E,\mathbf{W}_{\Upsilon}^E;\beta) &= \mathcal{L}_{t,P}(\mathbf{\Phi}_{\Delta}^E,\mathbf{\Phi}_{\Upsilon}^E,\mathbf{W}_{\Upsilon}^E) \\
    & + \mathcal{L}_{t,T}(\mathbf{\Phi}_{\Delta}^E,\mathbf{\Phi}_{\Gamma}^E,\mathbf{W}_\Gamma^E)\\
    & + \mathcal{L}_{t,W}(\mathbf{W}_\Delta^E,\mathbf{\Phi}_{\Delta}^E) \\
    & - \beta \mathcal{L}_{t,I}(\mathbf{\Phi}_{\Upsilon}^E)
\end{split}
\label{eq:overallLoss}
\end{equation}
where $\beta$ is obtained using hyperparameter tuning as described in section \ref{sec:implementation} and superscript $E$ stands for parameters pertaining to encoder.
Outcome representation, confounding representation and treatment at timestep $t$ is then used to forecast the outcome at the next timestep  $\hat{y}_{t+1}$ as described in Eq. \ref{eq:nextStepOutcome}.

\subsection{Decoder}
Our larger objective is to obtain $y_{t+\tau}$ (Eq. \ref{eq:estimand}) i.e,  forecast the outcome for a sequence of $\tau$ future treatments. To that end, we use a sequence-to-sequence architecture, which extends the next timestep ahead outcome prediction using our encoder to a $\tau$-step ahead prediction using a decoder, where an optimal encoder is learned first, then a decoder is updated as described in the text that follows.

We concat disentangled representations from the optimal encoder to obtain a unified representation $\mathbf{\Psi}_{t} = \mathbf{\Phi}_{\Upsilon}^E(\mathbf{s}_t) \oplus \mathbf{\Phi}_{\Delta}^E(\mathbf{s}_t)\oplus \mathbf{\Phi}_{\Gamma}^E(\mathbf{s}_t)$. Representation $\mathbf{\Psi}_t$ is obtained for each timestep for $N$ patients. Observational dataset is processed by splitting each patient's trajectory into shorter sequences of $\tau$ timesteps: $\{\{\mathbf{\Psi}_{t}\} \cup \{y_{t+m},\mathbf{a}_{t+m},y_{t+m+1}\}_{m=1}^{\tau-1} \cup \{\mathbf{v}\}\}_{t=1}^{T-\tau}$ for training the decoder.

The representation $\mathbf{\Psi}_t$ is used to initialize the state of the decoder recurrent network  and to finally obtain a state representation for the patient  $m$-timesteps in the future:
\begin{equation}
    \mathbf{s}_{t+m} = f_{m,s}^D([\mathbf{v},y_{t+m},\mathbf{a}_{t+m-1}])
    \label{eq:recurrentStateDec}
\end{equation}
where inputs to the LSTM unit of the decoder network are static covariates $\mathbf{v}$, previous timestep's outcome $y_{t+m}$ and previous treatment $\mathbf{a}_{t+m-1}$ with superscript $D$ pertaining to decoder parameters.
The state representation ($\mathbf{s}_{t+m}$) is further disentangled into confounding ($\mathbf{\Phi}_{\Delta}^D(\mathbf{s}_{t+m})$), treatment ($\mathbf{\Phi}_{\Gamma}^D(\mathbf{s}_{t+m})$), outcome ($\mathbf{\Phi}_{\Upsilon}^D(\mathbf{s}_{t+m})$) representations using $\mathcal{L}_{t+m}(\mathbf{\Phi}_{\Delta}^D,\mathbf{\Phi}_{\Upsilon}^D,\mathbf{\Phi}_{\Gamma}^D,\mathbf{W}_{\Delta}^D,\mathbf{W}_{\Gamma}^D,\mathbf{W}_{\Upsilon}^D;\beta)$, which is the same  loss as in Eq.~\ref{eq:overallLoss} computed for $m$ timesteps ahead of the $t^{th}$ timestep for learning decoder parameters. We then predict outcome $\hat{y}_{t+m+1}$ for $m+1$ timesteps ahead of $t$ as:
\begin{equation}
    \hat{y}_{t+m+1} = \mathbf{W}_{\Upsilon}^D([\mathbf{\Phi}_{\Upsilon}^D(\mathbf{s}_{t+m}),\mathbf{\Phi}_{\Delta}^D(\mathbf{s}_{t+m}),\mathbf{a}_{t+m}])
\end{equation}
and proceed iteratively for all values of $m$ upto $\tau-1$ timesteps ahead of $t$.

Figure \ref{fig:architecture} demonstrates sequence-to-sequence network architecture for next step ahead prediction and $\tau$-step ahead prediction. It should be noted that the true $y_{t+m}$ is not available as input during test time for the decoder. Hence, previous timestep's outcome predicted by decoder $\hat{y}_{t+m}$ are auto-regressively used as input to decoder's recurrent network (Eq. \ref{eq:recurrentStateDec}) for prediction of $\hat{y}_{t+m+1}$ (refer Figure \ref{fig:architecture}).




\vspace{-0.4em}
\section{Experimental Set-up}
\label{sec:setup}
In this section, we discuss the experimental set-up used for measuring the efficacy of the proposed model. First we describe the dataset, followed by baselines, evaluation metrics and implementation specifics for evaluation of DRTCI.
\subsection{Dataset}
As real world datasets lack counterfactual outcomes, we use a  bio-model \cite{geng2017prediction} for simulating a lung cancer dataset which contains the outcomes for different treatment options: no treatment, chemotherapy, radiotherapy, and both chemotherapy and radiotherapy treatments on tumour growth volume in a longitudinal manner. The generation model is the same as used by \cite{lim2018forecasting,bica2020} and we briefly describe the simulation model in Section \ref{sec:data_sim}.

We generate a dataset for a maximum of $60$ timesteps and evaluate DRTCI under different degrees of time-dependent confounding ($\gamma_c$,$\gamma_r$) and time horizons ($\tau$). For each setting of $\gamma_c,\gamma_r,\tau$, we simulate $10000$ patients for training, $1000$ for validation and $1000$ for testing. For testing of next step ahead prediction using an encoder, we simulate volume $V(t+1)$ for each timestep $t$ under all treatment options for each patient while for $\tau$-step ahead prediction using the decoder, we generate $2\tau$ counterfactual sequences of outcomes with each sequence giving chemotherapy at one of $t, ...,t+\tau-1$ and similarly radiotherapy at one of $t,..t+\tau-1$ timesteps for each timestep $t$ in the patient's trajectory. 

\subsection{Baselines}
\label{sec:baselines}
We benchmark DRTCI against 1) Marginal Structural Models (MSMs) \cite{robins2000marginal} which use logistic regression for inverse propensity weighting and linear regression for prediction, 2) Recurrent Marginal Structural Networks (RMSNs)  which use a sequence-to-sequence architecture for handling confounders and outcome prediction. \cite{lim2018forecasting} and  3) the  state of the art Counterfactual Recurrent Network (CRN) which introduces adversarial balancing for handling confounders and outcome prediction. 

\subsection{Evaluation Metric \& Implementation Specifics}
\label{sec:implementation}
We test the performance of DRTCI in terms of normalized root mean squared error \% (NRMSE \%) with NRMSE \% = (RMSE/maximum tumour volume)*100, where RMSE$ = (\frac{\sum_i\sum_t(\hat{y}_{t+1} - y_{t+1})^2}{NT})^{\frac{1}{2}}$ for encoder, with $y_{t+1}$ replaced by $y_{t+\tau}$ for decoder error computation. Details of implementation specifics are discussed in Section \ref{sec:hyperparameter}.




\section{Experimental Evaluations}
\label{sec:exptt}
In this section, we evaluate our proposed approach DRTCI and compare its performance with baselines for varying degrees of time-varying confounding and for $\tau$-step ahead counterfactual predictions.
\subsection{Comparison with Baselines}

\begin{table}[!b]
\footnotesize
    \centering
    \scalebox{0.8}{
    \begin{tabular}{ccccccc}
        \toprule
          &  &$\tau=3$ & $\tau = 4$ & $\tau = 5$ & $\tau = 6$ & $\tau = 7$  \\
         \toprule
            \multirow{1}{*}{(a)}& DRTCI & $\mathbf{2.01\%}$ & $\mathbf{2.66\%}$ & $\mathbf{3.00}\%$ & $\mathbf{2.81\%}$ & $\mathbf{2.14\%}$ \\
        & CRN & $2.43\%$ & $2.83\%$ & $3.18\%$ & $3.51\%$ & $3.93\%$ \\
          &RMSN & $3.16\%$ & $3.95\%$ & $4.37\%$ & $5.61\%$ & $6.21\%$ \\
          & MSM & $6.75\%$ & $7.65\%$ & $7.95\%$ & $8.19\%$ & $8.52\%$ \\
          \midrule
           
            \multirow{2}{*}{(b)}& DRTCI & $\mathbf{1.11\%}$ & $\mathbf{1.15\%}$ & $\mathbf{1.38}\%$ & $\mathbf{1.31\%}$ & $\mathbf{1.41\%}$ \\
          & CRN & $1.54\%$ & $1.81\%$ & $2.03\%$ & $2.23\%$ & $2.43\%$ \\
          & RMSN & $1.59\%$ & $2.25\%$ & $2.71\%$ & $2.73\%$ & $2.88\%$ \\
          & MSM & $3.23\%$ & $3.52\%$ & $3.63\%$ & $3.71\%$ & $3.79\%$ \\
          \midrule
       \multirow{3}{*}{(c)}& DRTCI & $1.31\%$ & $\mathbf{1.16\%}$ & $\mathbf{1.31\%}$ & $\mathbf{1.21\%}$ & $\mathbf{1.31\%}$ \\
          & CRN & $\mathbf{1.08\%}$ & $1.21\%$ & $1.33\%$ & $1.42\%$ & $1.53\%$ \\
            & RMSN & $1.35\%$ & $1.81\%$ & $2.13\%$ & $2.41\%$ & $2.43\%$ \\
          & MSM & $3.68\%$ & $3.84\%$ & $3.91\%$ & $3.97\%$ & $4.04\%$ \\
          \bottomrule
    \end{tabular}}
    \caption{NRMSE \% comparisons for (a) $\gamma_c = \gamma_r =5$, (b) $\gamma_r = 5, \gamma_c = 0$, (c) $\gamma_r = 0,\gamma_c =5$ for $\tau$-step ahead prediction.}
    \label{tab:baselines}
\end{table}
 %
We compare DRTCI for varying degrees of confounding: (a) $\gamma_c = \gamma_r =5$, (b) $\gamma_r = 5, \gamma_c = 0$, (c) $\gamma_r = 0,\gamma_c =5$ with the baselines approaches mentioned in Section \ref{sec:baselines}. We analyze this comparison for $\tau$-step ahead counterfactual prediction with $\tau$ increasing in steps of $1$ from $3$ to $7$ in Table \ref{tab:baselines}. DRTCI substantially outperforms RMSN, MSM and the current state of art CRN for all different values of $\tau$ and degree of confounding ($\gamma_r,\gamma_c$).

We further analyze the performance of DRTCI for very high degrees of confounding by setting $\gamma_r=\gamma_c=\gamma \in \{6,7,8,9,10\}$ for $\tau=1$ (next step ahead prediction) and $\tau = 3,5$ in Table \ref{tab:degree}.   Results show DRTCI performance for high confounding is  superior than CRN for next step ahead prediction and better by large margins for $\tau$-step ahead prediction achieving more than 32\% and 25\% reduction in errors for $\tau=3$ and $\tau=5$ respectively. We believe this is due to the fact that  DRTCI learns confounding representation which plays a significant role in outcome prediction especially when the degree of confounding is high, which is in contrast to prior approaches where confounding covariates are balanced for reduction in bias and reliable outcome prediction. Moreover, while in prior approaches a shared outcome regression arm would lead to noisy interference between the treatment predictions, in DRTCI since only the relevant factors are being used to predict outcomes, noisy interference is greatly reduced and common knowledge is better utilized across treatments. We also test the quality of treatment sequence prediction $p_2(s_t)$ in Section \ref{sec:appx_treatment_prediction} for varying degrees of confounding.

\begin{table}[]
\footnotesize
    \centering
    \addtolength{\tabcolsep}{-1pt}
    \scalebox{0.8}{
    \begin{tabular}{c|cc|cc|cc}
    \toprule
         & \multicolumn{2}{c}{$\tau = 1$}  &\multicolumn{2}{c}{$\tau = 3$}  &\multicolumn{2}{c}{$\tau = 5$}  \\
         \midrule
        $\gamma$ & DRTCI&CRN &DRTCI &CRN &DRTCI &CRN  \\
        \midrule
        6 &$\mathbf{1.54\%}$ & $1.72\%$ &$\mathbf{2.87}\%$ & $4.64\%$ & $\mathbf{2.87\%}$& $4.26\%$\\
        7 & $\mathbf{1.87\%}$ & $1.89\%$ &$\mathbf{3.82\%}$ & $4.65\%$ & $\mathbf{3.66\%}$& $5.81\%$\\
        8 &$\mathbf{2.60\%}$ & $2.78\%$ & $\mathbf{4.99\%}$& $8.13\%$ & $\mathbf{4.65\%}$&$7.72\%$\\
        9 & $\mathbf{3.03\%}$ & $3.06\%$ & $\mathbf{6.57\%}$& $8.32\%$ & $6.55\%$& $\mathbf{6.24\%}$\\
        10& $\mathbf{4.17\%}$ & $4.26\%$ & $\mathbf{6.66\%}$& $11.32\%$ & $\mathbf{6.51\%}$& $8.53\%$\\
         Mean & $\mathbf{2.64\%}$ & $2.74\%$ & $\mathbf{4.98\%}$ & $7.41\%$ & $\mathbf{4.85\%}$ & $6.51\%$\\
        \bottomrule
    \end{tabular}}
    \caption{NRMSE \% of DRTCI for high $\gamma$ (high confounding)}
    \label{tab:degree}
\end{table}
\subsection{Ablation Study}
An extensive ablation study was carried out to identify the key factors contributing towards the performance of DRTCI. We perform this study for $5$-step ahead prediction for varying degree of confounding with $\gamma=3,7,10$. We examine the following conditions: (a) $\mathbf{\Phi}_{\Upsilon}$: DRTCI with no disentangled representation but only $\mathbf{\Phi}_{\Upsilon}$ as the representation layer, (b) $\mathbf{\Phi}_{\Upsilon} \oplus \mathbf{\Phi}_{\Delta}$: DRTCI with outcome representation ($\mathbf{\Phi}_{\Upsilon}(\cdot)$), confounding representation ($\mathbf{\Phi}_{\Delta}(\cdot)$) and no separate representation for only treatment ($\mathbf{\Phi}_{\Gamma}(\cdot)$), (c) $\mathbf{\Phi}_{\Upsilon} \oplus \mathbf{\Phi}_{\Gamma}$: DRTCI with outcome representation ($\mathbf{\Phi}_{\Upsilon}(\cdot)$), treatment representation ($\mathbf{\Phi}_{\Gamma}(\cdot)$) and no representation for confounding ($\mathbf{\Phi}_{\Delta}(\cdot)$) in Table \ref{tab:ablation}.



Analyzing the representation layer which contributes most towards the performance, we see that the outcome representation (column $\mathbf{\Phi}_{\Upsilon}$) alone does not provide significant performance gains while learning treatment or confounding representation along with outcome representation ($\mathbf{\Phi}_{\Upsilon} \oplus \mathbf{\Phi}_{\Delta},\mathbf{\Phi}_{\Upsilon} \oplus \mathbf{\Phi}_{\Gamma}$) gives a significant performance gain. It can be safely concluded that using the proposed representations which retain confounding and treatment factors such as in DRTCI, has the most significant impact on improvement of counterfactual prediction in longitudinal data, especially when the degree of confounding is high.

    %
        
 %
\vspace{-1em}
\begin{table}[!h]
\footnotesize
    \centering
    \addtolength{\tabcolsep}{-1pt}
    \scalebox{0.8}{
    \begin{tabular}{ccccccc}
         \midrule
         $\gamma$ & DRTCI & $\mathbf{\Phi}_{\Upsilon}$ & $\mathbf{\Phi}_{\Upsilon} \oplus \mathbf{\Phi}_{\Delta}$ & $\mathbf{\Phi}_{\Upsilon} \oplus \mathbf{\Phi}_{\Gamma}$ & CRN\\
         
         \midrule
         $3$ & $1.58\%$ & $1.76\%$ & $\mathbf{1.56\%}$ & $\mathbf{1.56\%}$ & $1.72\%$\\
         $7$ & $\mathbf{3.66\%}$ & $5.02\%$ & $4.19\%$ & $4.65\%$ & $5.81\%$\\
         $10$ & $\mathbf{6.51\%}$ & $7.87\%$ & $6.97\%$ & $6.61\%$ & $8.53\%$\\
        
         \bottomrule
         
    \end{tabular}}
    \caption{Ablation Study for DRTCI $\tau=5$ step ahead performance}
    \label{tab:ablation}
\end{table}
\subsection{Error vis-a-vis length of history} 
\begin{figure}
  \centering
  \includegraphics[scale=0.23]{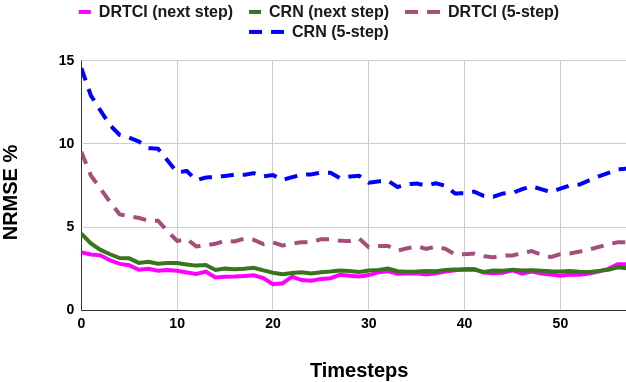}
  \caption{NRMSE \% versus history sizes for $\gamma=8$}
\label{fig:historyplot}
\end{figure}

\begin{figure}
  \centering
  \includegraphics[scale=0.2]{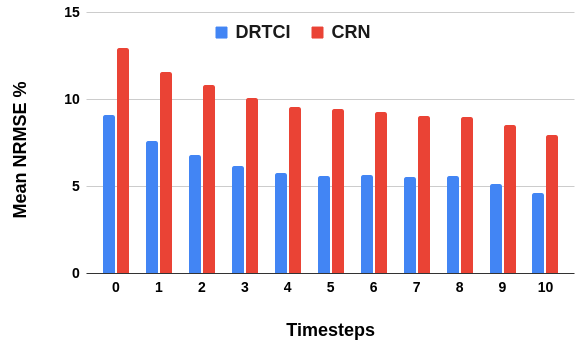}
  \caption{Early prediction mean NRMSE \% across $\gamma$ for $\tau=5$}
\label{fig:meanhistplot}
\end{figure}

Medical decisions made early in the 'golden window' of patient admission can be critical to the final outcome, therefore it is important to evaluate how well our model performs with limited historical information for early treatment prediction.
We evaluate the performance of DRTCI for next step ahead and $5$-step ahead prediction for varying history sizes. We depict our analysis for high confounding with $\gamma=8$ in Figure \ref{fig:historyplot}. It is seen that for next step prediction, DRTCI performs marginally better than CRN with minimal history, while for $5$-step prediction DRTCI performs significantly better for all history sizes. We also analyze DRTCI for different history lengths in Figure \ref{fig:meanhistplot} using mean of NRMSE\% values computed for high confounding. It is seen that DRTCI significantly outperform CRN for early prediction. Thus, when making distant early predictions in high confounding cases, DRTCI is more reliable.
\vspace{-0.4em}
\section{Conclusion and Future Work}
While many real world applications for causal inference  involve temporal observational data, the literature has largely focused  on techniques for the static data setting. Recent work has harnessed the power of representation learning to mitigate selection bias during counterfactual estimation, and has demonstrated the value of disentangling representations for counterfactual estimation in the static data setting. This paper presents DRTCI, a model that extends this disentanglement to the temporal setting and demonstrates significant benefits over the current state of the art for treatment prediction in a realistic medical scenario, especially for early predictions over long time horizons in high confounding situations. Future work involves extending this work for high cardinality treatments and to employ meta learning techniques to reduce data requirements and enhance adaptability to novel datasets similar to \cite{sharma2019metaci}.

\label{sec:conclusion}

\bibliography{example_paper}

\begin{thebibliography}{18}
\providecommand{\natexlab}[1]{#1}
\providecommand{\url}[1]{\texttt{#1}}
\expandafter\ifx\csname urlstyle\endcsname\relax
  \providecommand{\doi}[1]{doi: #1}\else
  \providecommand{\doi}{doi: \begingroup \urlstyle{rm}\Url}\fi

\bibitem[Ankit et~al.(2019)Ankit, Garima, Ranjitha, Arnab, Lovekesh, and
  Gautam]{sharma2019metaci}
Ankit, S., Garima, G., Ranjitha, P., Arnab, C., Lovekesh, V., and Gautam, S.
\newblock Metaci: Meta-learning for causal inference in a heterogeneous
  population.
\newblock \emph{arXiv preprint arXiv:1912.03960}, 2019.

\bibitem[Ankit et~al.(2020)Ankit, Garima, Ranjitha, Arnab, Lovekesh, and
  Gautam]{sharma2020hi}
Ankit, S., Garima, G., Ranjitha, P., Arnab, C., Lovekesh, V., and Gautam, S.
\newblock Hi-ci: Deep causal inference in high dimensions.
\newblock In \emph{Proceedings of the 2020 KDD Workshop on Causal Discovery},
  pp.\  39--61. PMLR, 2020.

\bibitem[Austin(2011)]{austin2011introduction}
Austin, P.~C.
\newblock An introduction to propensity score methods for reducing the effects
  of confounding in observational studies.
\newblock \emph{Multivariate behavioral research}, 46\penalty0 (3):\penalty0
  399--424, 2011.

\bibitem[Bica et~al.(2020)Bica, Alaa, Jordon, and van~der Schaar]{bica2020}
Bica, I., Alaa, A.~M., Jordon, J., and van~der Schaar, M.
\newblock Estimating counterfactual treatment outcomes over time through
  adversarially balanced representations, 2020.

\bibitem[Bottou et~al.(2013)Bottou, Peters, Qui{\~n}onero-Candela, Charles,
  Chickering, Portugaly, Ray, Simard, and Snelson]{bottou2013counterfactual}
Bottou, L., Peters, J., Qui{\~n}onero-Candela, J., Charles, D.~X., Chickering,
  D.~M., Portugaly, E., Ray, D., Simard, P., and Snelson, E.
\newblock Counterfactual reasoning and learning systems: The example of
  computational advertising.
\newblock \emph{The Journal of Machine Learning Research}, 14\penalty0
  (1):\penalty0 3207--3260, 2013.

\bibitem[Cole et~al.(2010)Cole, Platt, Schisterman, Chu, Westreich, Richardson,
  and Poole]{cole2010illustrating}
Cole, S.~R., Platt, R.~W., Schisterman, E.~F., Chu, H., Westreich, D.,
  Richardson, D., and Poole, C.
\newblock Illustrating bias due to conditioning on a collider.
\newblock \emph{International journal of epidemiology}, 39\penalty0
  (2):\penalty0 417--420, 2010.

\bibitem[Geng et~al.(2017)Geng, Paganetti, and Grassberger]{geng2017prediction}
Geng, C., Paganetti, H., and Grassberger, C.
\newblock Prediction of treatment response for combined chemo-and radiation
  therapy for non-small cell lung cancer patients using a bio-mathematical
  model.
\newblock \emph{Scientific reports}, 7\penalty0 (1):\penalty0 1--12, 2017.

\bibitem[Hassanpour \& Greiner(2019)Hassanpour and
  Greiner]{hassanpour2019learning}
Hassanpour, N. and Greiner, R.
\newblock Learning disentangled representations for counterfactual regression.
\newblock In \emph{International Conference on Learning Representations}, 2019.

\bibitem[Johansson et~al.(2016)Johansson, Shalit, and
  Sontag]{johansson2016learning}
Johansson, F., Shalit, U., and Sontag, D.
\newblock Learning representations for counterfactual inference.
\newblock In \emph{International conference on machine learning}, pp.\
  3020--3029, 2016.

\bibitem[Lim(2018)]{lim2018forecasting}
Lim, B.
\newblock Forecasting treatment responses over time using recurrent marginal
  structural networks.
\newblock In \emph{Advances in Neural Information Processing Systems}, pp.\
  7483--7493, 2018.

\bibitem[Mansournia et~al.(2017)Mansournia, Etminan, Danaei, Kaufman, and
  Collins]{mansournia2017handling}
Mansournia, M.~A., Etminan, M., Danaei, G., Kaufman, J.~S., and Collins, G.
\newblock Handling time varying confounding in observational research.
\newblock \emph{Bmj}, 359, 2017.

\bibitem[Negar \& Russell(2019)Negar and Russell]{hassanpour2019counterfactual}
Negar, H. and Russell, G.
\newblock Counterfactual regression with importance sampling weights.
\newblock In \emph{IJCAI}, pp.\  5880--5887, 2019.

\bibitem[Pearl \& Mackenzie(2018)Pearl and Mackenzie]{pearl2018book}
Pearl, J. and Mackenzie, D.
\newblock \emph{The book of why: the new science of cause and effect}.
\newblock Basic Books, 2018.

\bibitem[Robins et~al.(2000)Robins, Hernan, and Brumback]{robins2000marginal}
Robins, J.~M., Hernan, M.~A., and Brumback, B.
\newblock Marginal structural models and causal inference in epidemiology,
  2000.

\bibitem[Shalit et~al.(2017)Shalit, Johansson, and
  Sontag]{shalit2017estimating}
Shalit, U., Johansson, F.~D., and Sontag, D.
\newblock Estimating individual treatment effect: generalization bounds and
  algorithms.
\newblock In \emph{International Conference on Machine Learning}, pp.\
  3076--3085. PMLR, 2017.

\bibitem[Sharma et~al.(2020)Sharma, Gupta, Prasad, Chatterjee, Vig, and
  Shroff]{sharma2020multimbnn}
Sharma, A., Gupta, G., Prasad, R., Chatterjee, A., Vig, L., and Shroff, G.
\newblock Multimbnn: Matched and balanced causal inference with neural
  networks.
\newblock \emph{ESAAN}, 2020.

\bibitem[Swaminathan \& Joachims(2015)Swaminathan and
  Joachims]{swaminathan2015self}
Swaminathan, A. and Joachims, T.
\newblock The self-normalized estimator for counterfactual learning.
\newblock \emph{advances in neural information processing systems},
  28:\penalty0 3231--3239, 2015.

\bibitem[Yao et~al.(2018)Yao, Li, Li, Huai, Gao, and
  Zhang]{yao2018representation}
Yao, L., Li, S., Li, Y., Huai, M., Gao, J., and Zhang, A.
\newblock Representation learning for treatment effect estimation from
  observational data.
\newblock \emph{Advances in Neural Information Processing Systems},
  31:\penalty0 2633--2643, 2018.

\end{thebibliography}
\bibliographystyle{icml2021}

\appendix

\section{Implementation Specifics}


\label{sec:hyperparameter}
\begin{table}[!h]
    \footnotesize
    \centering
    \scalebox{0.8}{ \begin{tabular}{ccc}
    \toprule
         Hyperparameter & Encoder & Decoder  \\
         \midrule
         Epochs & 100 & 50\\
         Batch Size & 256 & 1024\\
         Learning rate & 0.001&0.001 \\
         MLP hidden units &50,100&50,100\\
         Recurrent hidden units & 50,100,150 & Size of $\mathbf{\Psi}_t$\\
         Dropout & 0.1 & 0.1\\
         $\beta$ & 0.1,0.4,0.7,1 & 0.1,0.4,0.7,1\\
         \bottomrule
    \end{tabular}}
    \caption{Encoder and decoder hyperparameter search space}
    \label{tab:hyperparams}
\end{table}

Table \ref{tab:hyperparams} illustrates the search range of hyperparameters for encoder and decoder networks in DRTCI. 
From the search space of hyperparameters, we select optimal hyperparameters based on minimum NRMSE\% in predicting the factual outcome for the validation dataset.  

For evaluation of NRMSE \%, the maximum tumour volume in the data is $1150$cm${}^3$

We use a Tesla v100 with 4GB GPU, 32 GB RAM for training and hyperparameter optimization for our experiments.  

\section{Treatment prediction}
\label{sec:appx_treatment_prediction}
We test treatment sequence prediction on $1000$ patients with treatment sampled from $p_2(\mathbf{s}_t)$, where $\hat{\mathbf{a}_t} \sim p_2(\mathbf{s}_t)$ and evaluated using accuracy as metric for different degrees of confounding $\gamma$. It is seen for increasing $\gamma$, accuracy of treatment prediction increases. This is because with increase in degree of confounding, factual treatment assignment is more biased and thus, is easier to predict.

\begin{table}[!h]
    \centering
    \scalebox{0.8}{ \begin{tabular}{cc}
    \toprule
         $\gamma$ & Accuracy  \\
         \midrule
         0 & $62\%$\\
         1 & $67.27\%$\\
         2 & $71.82\%$\\
         3 & $75.59\%$\\
         4 & $78.19\%$\\
         5 & $79.90\%$\\
         6 & $81.07\%$\\
         7 & $82.33\%$\\
         8 & $82.64\%$\\
         9 & $83.68\%$\\
         10 & $83.90\%$\\
         
         \bottomrule
    \end{tabular}}
    \caption{Accuracy of treatment sequence prediction}
    \label{tab:p_2}
\end{table}

\section{Data Simulation}
\label{sec:data_sim}
To begin with, initial cancer stage and initial tumour volume is obtained from prior distributions for each patient. The volume of the tumour $t$ days after diagnosis is then obtained using the following mathematical model:
\begin{equation}
    \resizebox{0.99\hsize}{!}{$V(t+1) = (1 + \rho log(\frac{\kappa}{V(t)}) - \beta_cC(t) - (\alpha_rd(t)+\beta_rd(t)^2) + e(t))V(t))$}
    \label{eq:datasimulate}
\end{equation}
where $\kappa$,$\rho$, $\beta_c$, $\alpha_r$, $\beta_r$, $e_t$ are sampled as described in \cite{geng2017prediction}. Time-varying confounding is introduced by modelling chemotherapy and radiotherapy assignment as Bernoulli random variables with probability of chemotherapy ($p_c$) and radiotherapy ($p_r$) depending upon tumour volume: $p_c(t) = \sigma( \frac{\gamma_c}{D_{max}}(\bar{D}(t) - \delta_c))$ and $p_r(t) = \sigma(\frac{\gamma_r}{D_{max}}(\bar{D}(t) - \delta_r))$, where $\gamma_c$, $\gamma_r$ are the factors controlling degree of time-varying confounding, $\bar{D}_t$ is the average diameter over the last $15$ days, $D_{max} = 13$ cm, $\sigma(\cdot)$ is the sigmoid and $\delta_r = \delta_c = D_{max}/2$. Chemotherapy concentration is given by $C(t) = 5 + C(t-1)/2$ and radiotherapy dose is $d(t) = 2$ if applied at timestep $t$. The implementation of data simulator can be obtained from git repository \footnote{\url{https://github.com/ioanabica/Counterfactual-Recurrent-Network/blob/master/utils/cancer_simulation.py}}. 

It is to be noted that inspite that the representation size of DRTCI is three times that of baseline such as CRN, we simulate data of same size as that in \cite{bica2020} to resemble real-like data and for fare comparisons.

\end{document}